\documentclass{article}
\usepackage{spconf,amsmath,graphicx}
\usepackage{amsmath}
\usepackage{algorithm}
\usepackage[noend]{algpseudocode}
\usepackage{multirow}
\usepackage[inline]{enumitem}
\usepackage{subfig}
\usepackage{color}
\usepackage[usenames,dvipsnames,svgnames,table]{xcolor}
\usepackage{graphicx}
\usepackage{fancyhdr}
\usepackage[inline]{enumitem}
\usepackage{url}
\usepackage{subfig}
\usepackage{amsmath}
\usepackage{colortbl}
\usepackage{multirow}

\title{Wearable Audio and IMU Based Shot Detection in Racquet Sports}
\name{Manish Sharma, Akash Anand, Rupika Srivastava and Lakshmi Kaligounder}
\address{Samsung R\&D Institute India, Bangalore}

\begin{document}
%
\maketitle
\begin{abstract}
Wearables like smartwatches which are embedded with sensors and powerful processors, provide a strong platform for development of analytics solutions in sports domain. To analyze players' games, while motion sensor based shot detection has been extensively studied in sports like Tennis, Golf, Baseball; Table Tennis and Badminton are relatively less explored due to possible less intense hand motion during shots. In our paper, we propose a novel, computationally inexpensive and real-time system for shot detection in table tennis, based on fusion of Inertial Measurement Unit (IMU) and audio sensor data embedded in a wrist-worn wearable. The system builds upon our presented methodology for synchronizing IMU and audio sensor input in time using detected shots and achieves 95.6\% accuracy. To our knowledge, it is the first fusion-based solution for sports analysis in wearables. Shot detectors for other racquet sports as well as
further analytics to provide features like shot classification, rally analysis and recommendations, can easily be built over our proposed solution.

\end{abstract}
\section{Introduction}
\label{sec:intro}

In the health and fitness domain, some available smart watch based solutions (like Samsung Health, Pebble Sleep) and standalone devices (like Zepp Tennis swing analyzer, Coollang badminton) use sensors to aid users in general activity and sports tracking. Majority of these solutions and the research pursued in this field utilize data from the embedded Inertial Measurement Units (IMUs) in devices. However, using microphones in conjunction with IMU, particularly in the sports field, can improve their analysis and address drawbacks of solutions that use only IMU sensors. In this paper, we propose a system to analyze swing-based sports like Table Tennis, Badminton etc. by utilizing motion data from IMU and audio from microphone, each embedded in a wrist-worn wearable. The audio captured during racquet impact, IMU data representing hand motion of a player and a proposed methodology to synchronize data from these two different sensor sources form the basis of our generalized algorithm for shot detection in such sports. As a case study, the efficacy of the system in detecting shots in Table Tennis is presented, which is otherwise difficult using IMU or microphone data alone. In our understanding, there has been no research on combining the data from microphone and IMU to provide analysis for sports in wearables. 

\vspace{-8pt}
\section{RELATED WORK}
\label{sec:prior}
Researchers have primarily adopted IMU based or audio-visual based approaches for sports analysis. Among IMU based approaches, previous works have attempted shot analysis using hardware on either the playing equipment or on the forearm. In \cite{pei2017embedded}, Pie et al. presented an embedded device in tennis racket handle to detect tennis shots with an accuracy of 98\%. In other work, Srivastava et al. \cite{srivastava2015efficient} and Connaghan et al.\cite{connaghan2011multi} proposed algorithms for shot detection in tennis, using a wrist worn equipment with accuracies of  98\% and 90\% respectively. The audio-visual based research work in sports use one or multiple recorders placed at predefined locations in the playing area. Yoshikawa et al \cite{yoshikawa2010automated} discussed an algorithm for service detection in sports using a ceiling camera, with an accuracy of 95\% for badminton serve. A popular usecase of audio based sports analysis has been to generate match highlights \cite{wang2004sports}, \cite{zhang2006ball}, \cite{xiong2003audio}, \cite{rui2000automatically}. Xiong et al \cite{xiong2003audio} proposed MPEG-7 audio features based golf ball hit detection and used it for highlights generation. In \cite{zhang2006ball} Zhang et al. investigated table tennis shot detection and proposed Energy Peak Detection based lightweight approach that achieves F-score of 81\%. 

As compared to other popular sports like Tennis and Golf, Table Tennis and Badminton have received little attention of the IMU based research community. This is because flexibility of rotation about wrist allows more variation in shots, compared to Tennis and Golf. While past studies have quoted excellent results, our work is unique because it is wrist-wearable based, easily extensible to other sports, performs real-time tracking and capitalizes on a new feature space by combining audio with IMU data, that was not considered earlier.

\vspace{-10pt}
\section{System Diagram} 
\begin{figure}[t!]
\centering
	{\includegraphics[width=0.5\textwidth]{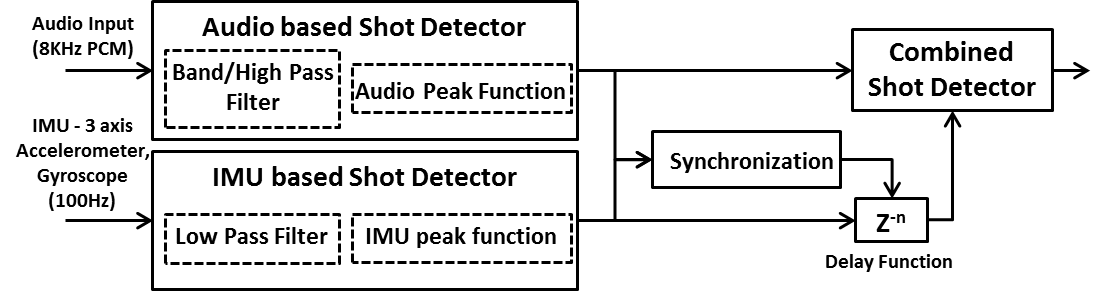}}
\caption{System diagram}
\label{fig:block_diagram}
\vspace{-12pt}
\end{figure}

The proposed system is shown in Fig. \ref{fig:block_diagram}. The input to system is audio from microphone and data from 3-axis accelerometer and gyroscope sensors (IMU) in the wrist-worn wearable.
There are four major blocks in the system. First, audio collected at 8KHz is input to the Audio based shot detector and second, IMU based shot detector takes inputs from IMU at 100Hz. Each generates a shot likelihood sequence at 100Hz. Third, the Synchronization block uses these likelihood sequences and computes the time offset between the two input modes. The calculated signal offset is validated for few seconds, and then Synchronization block is cut off. The audio and IMU data is synchronized using this offset and then used by the fourth and final block to give detection output. The next four sections elaborate on each of these blocks.

\vspace{-10pt}

\vspace{-0pt}
\section{Audio based Shot Detector}
\label{sec:aud}
Audio is a critical characteristic of shots, as the impact sets up pressure waves in the audible range which last for a short time (10-20ms for table tennis \cite{manin2012vibro}) before dying quickly. In \cite{zhang2006ball}, Zhang et al. propose a cascade of Energy Peak Detection (EPD) and MFCC-based Refinement (MBR) blocks to detect these transient variations in audio signal. EPD block captures coarse level events based on temporal variation and is followed by MBR block that detects fine level events based on C-SVC classification of frequency composition of signal. To develop our Audio based shot detector, we improve upon the EPD block from \cite{zhang2006ball}. In this paper, we substitute the nomenclature of EPD with Audio Peak Function (APF) for the ease of referencing. The details of APF (reproduced here for convenience) and filter sub-blocks are explained below.

\vspace{-10pt}
\subsection{Audio Peak Function (APF)}
$Microframe$ is defined as the set of samples in a window for which acoustic signal near impact typically persists. For table tennis, impact signal persists for about 10ms, so length of microframe is 10ms. $Macroframe$ is defined as the set of samples in a window for which most of the ambient noises can be considered stationary. Because short speeches last for atleast 100 ms, length of macroframe is fixed to 100ms. The incoming audio stream is divided into microframes, and for each microframe, the Short Time Energy (STE), $E$, is computed. $E$, given in Eq.\ref{Eq:APF},  represents the net signal energy in a microframe. Thereafter, APF which incorporates the context of a macroframe, is calculated for each microframe as given in Eq.\ref{EqnAPF}. APF is a time series at 100 Hz and exhibits high magnitude peaks at impact points. Thus, APF signal is considered as a measure of likelihood of shot impact in a microframe. 
\vspace{-5pt}
\begin{equation}
\begin{aligned}
E[i] = \sum_{s_k \epsilon     i^{th} microframe} s_k^2 \hspace{40pt} \label{Eq:APF}\\
\text{where } i = Microframe\text{ index};\\
\text{and } s_k = \text{ signal value at }k^{th}\text{ index in } microframe.\\
\end{aligned}
\end{equation}

\begin{equation}
\begin{aligned}
APF[i] = E[i] - \frac{1}{11} \sum_{j=i-5}^{i+5} E[j]\\ \label{EqnAPF}
\text{where } i, j = Microframe\text{ indices.}
\end{aligned}
\end{equation}

\vspace{-10pt}
\subsection{Audio based shot detector with filter and APF}

\begin{figure}[t]
	\centering
	{\includegraphics[width=0.5\textwidth]{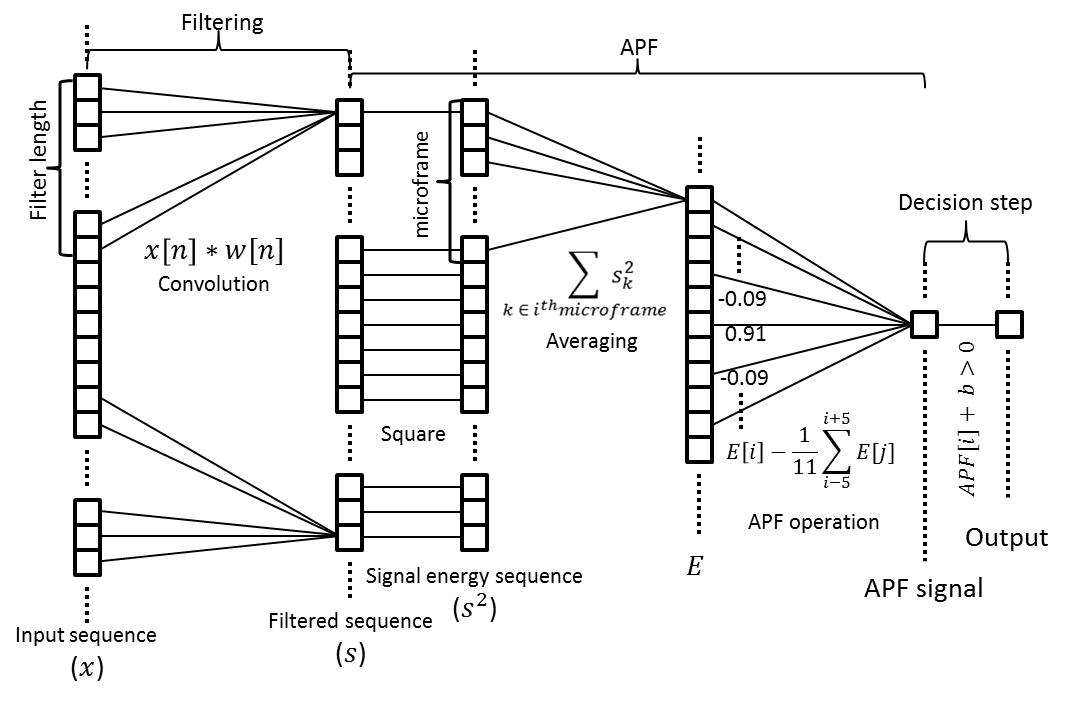}}
	\caption{Computational graph for the audio based shot detector}
	\label{fig:computational_graph}
	\vspace{-10pt}
\end{figure}

To improve performance of the APF sub-block, a filter sub-block is added before it to separate the shot sound based on its spectral signature. During the training phase, filter weights are determined by backpropagating classification loss of input shot and non-shot data, through the entire audio detector pipeline. In order to perform back propagation, the Audio based shot detector is represented as a computational graph shown in Fig. \ref{fig:computational_graph}. It consists of three steps, namely, filtering using FIR filter, APF computation stage and decision step. The first layer, similar to convolution layer in neural networks, filters the input($x$) signal sequence using convolution with filter weights($w$) of length $23$. Obtained filtered sequence ($s$) is used to calculate $E$ (Eq.\ref{Eq:APF}) in a microframe (80 samples) in the subsequent two layers. The resulting APF signal is obtained in the fourth layer using Eq. \ref{EqnAPF}. The decision step adds a bias to APF signal which mimics a thresholding operation and shot is declared when $APF[i] + bias > 0$ returns true. The FIR filter weights and the $bias$ term are obtained using stochastic gradient descent during training. Let true class label sequence be $y$ and predicted class label sequence be $\hat{y}$. The loss function, where $i$ corresponds to the $i^{th}$ microframe, is defined as:
\begin{align}
loss[i]=\left\{
	\begin{array}{ll}
		-(APF[i] + bias);\text{ if }y=1 \ \& \ \hat{y}=0\\
		(APF[i] + bias);\text{ if }y=0 \ \& \ \hat{y}=1\\
		0; \ \text{otherwise}
	\end{array}
\right.
\end{align}
The filter weights are initialized by random sampling from a zero mean normal distribution with appropriate variance whereas the bias term is initialized with zero. Audio signal sequences of both the shot and the non-shot instances are sampled in the ratio of 1:20 to account for the abundance of non-shots in a game. The network is trained in mini-batches with Adam optimizer~\cite{kingma2014adam} until the weights converge.

\vspace{-5pt}
\section{IMU based Shot Detector}
\label{sec:imu}

The IMU embedded in wrist-worn wearable captures the acceleration and angular velocity of the forearm during shots in swing based sports. This data has been used for shot detection in \cite{srivastava2015efficient} and \cite{connaghan2011multi}. In \cite{srivastava2015efficient}, Srivastava et al proposed an algorithm for tennis shot detection using Pan Tompkins on accelerometer data. However, it does not perform well for sports like table tennis, where the forearm acceleration during shots can be as low as acceleration due to gravity, making shots less separable from general hand motion. Connaghan et al \cite{connaghan2011multi} also discussed a method for shot detection building on the fact that acceleration magnitude during tennis strokes is more than $3g$. Around 62\% of the table tennis shots in our data set have an acceleration magnitude $<3g$ and consequently, using their approach is unsuitable. However the results in \cite{connaghan2011multi} showed that fusion of accelerometer and gyroscope provides better accuracy for shot detection than using accelerometer only. Hence, we use both accelerometer and gyroscope data to generate shot likelihood in IMU based shot detector. The constraints of using only IMU data are overcome by combining these results with cues from Audio based shot detector.

\subsection{Signal Preprocessing and IMU Peak Function (IPF)}
\begin{figure}[t]
	\centering
	\subfloat[]{\includegraphics[trim={0 0cm 0cm 0cm},clip,width=4.0cm]{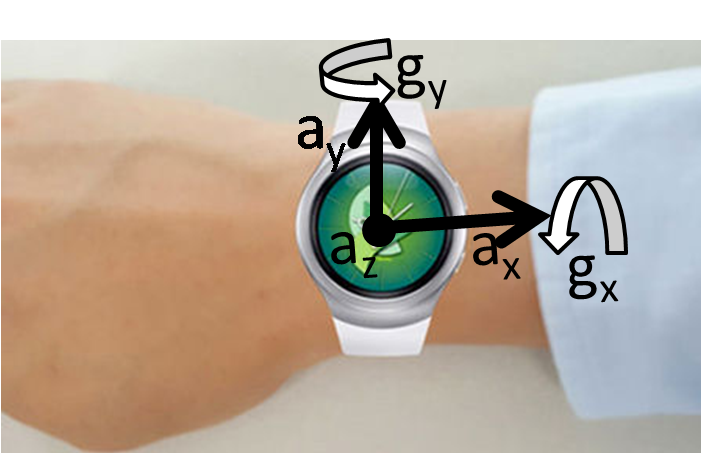}\label{image_gear_axes}}
	\quad
	\subfloat[]{\includegraphics[width=4.0cm]{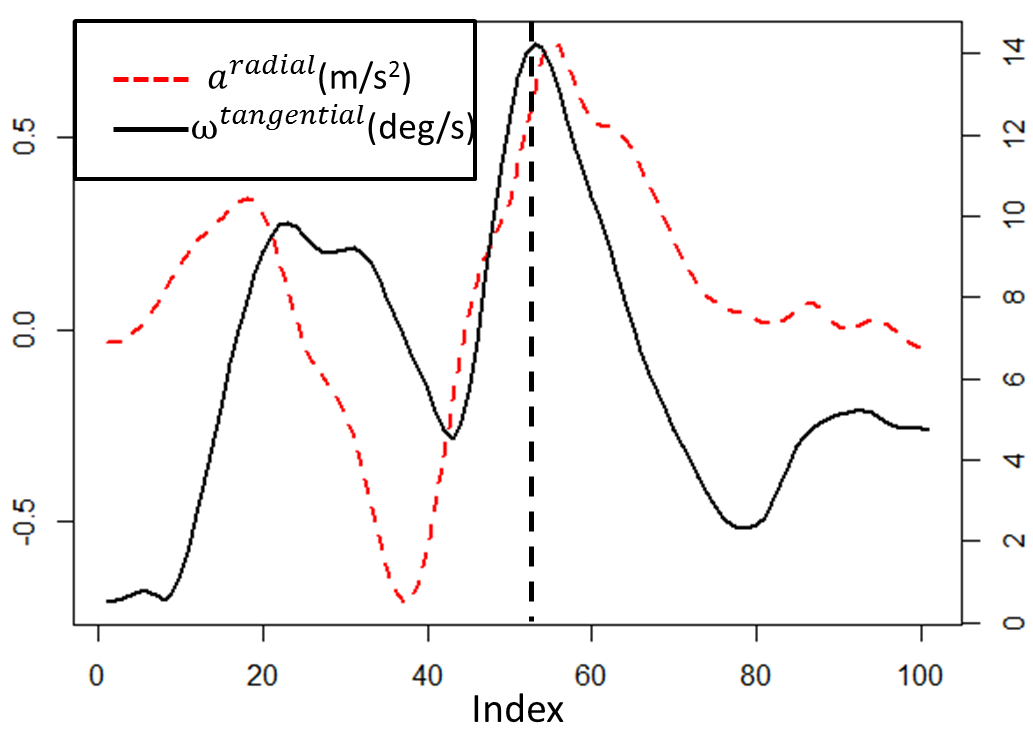}\label{image_sensor_data}}
	\quad
	\caption{\protect\subref{image_gear_axes} Direction of axes in smartwatch and \protect\subref{image_sensor_data} plots of radial acceleration and tangential angular velocity}
	\label{fig:IMU}
\end{figure}

Fig. \ref{fig:IMU} shows the wearable used in our study, along with the $x$, $y$ and $z- axes$ for accelerometer and gyroscope. The swing motion can be approximated as a circular motion with varying speed and radius. As the $x-axis$ is along the forearm, $a_x$ captures a significant component of the radial acceleration ($a_{rad}$ in Eq. \ref{eq:acceleration_components}) during such motion, hence, is characterized by a local maxima or peak near impact . The tangential component of acceleration is captured in $a_{tan}$. The angular velocity can also be divided into two components as given in Eq. \ref{eq:omega_components}, where tangential angular velocity $\omega_{tan}$ shows a maxima due to rotation of the arm during shots.

\vspace{-12pt}
\begin{align}
a_{rad} &= a_x  \text{,} \qquad a_{tan} = \sqrt{a_y^2 + a_z^2} \label{eq:acceleration_components} \\
\omega_{rad} &= \omega_x \text{,} \qquad \omega_{tan} = \sqrt{\omega_y^2 + \omega_z^2} \label{eq:omega_components}
\end{align}
\vspace{-8pt}

An IIR low-pass filter with cutoff frequency at 10Hz is applied to $a_{rad}$ and $\omega_{tan}$ signals to reduce signal noise. Fig. \ref{fig:IMU} shows local maxima near impact in these low passed signals. IMU Peak Function (IPF), given in Eq. \ref{EqnIPF}, is defined on similar lines as APF and uses these low-passed signals with a macroframe size of 10 samples (100ms) for each term in the product and produces peaks at shot impact points.

\vspace{-12pt}
\begin{align}
IPF[i] = \bigg(a_{x}[i] -  \sum_{j=i-4}^{i+5}a_{x}[j]\bigg)\times\bigg(\omega_{tan}[i] - \sum_{j=i-4}^{i+5}\omega_{tan}[j]\bigg) 
\label{EqnIPF}
\end{align}

\vspace{-0pt}
\section{Synchronizing Audio and IMU streams}

In order to collect data from IMU and microphone sensors, we register listeners for accelerometer, gyroscope and microphone on the wearable platform, and receive handler objects created by the OS to handle their data. As the process of object creation leads to a variable and unknown offset between data received from the two sources (typically few 100ms), it is imperative to synchronize the data before using them in conjunction for shot detection. We utilize the APF and IPF functions defined in Eq. \ref{EqnAPF} and \ref{EqnIPF} for this purpose.

It is already seen that APF and IPF peak values correspond to the likelihood of shot, and ideally their peaks should have perfectly aligned as the point of highest arm speed and that of audio impulse should be same. However, due to the audio and IMU data being asynchronously acquired and slight deviations in the location of peak values around actual impact points, a perfect synchronization is very difficult. The steps followed to estimate a near precise signal offset value are discussed below.

\begin{description}
\item [Step 1] Based on the prior probabilities $p(APF|shot)$ and $p(IPF|shot)$, 5 quintiles for APF and IPF are created and the respective values are transformed to these quantized levels.
\item [Step 2] Triangle smoothing is performed on APF and IPF signals by convolving with triangle waveform (1,2,3,4,3,2,1) resulting into $APF_{smooth}$ and $IPF_{smooth}$.
\item [Step 3] The offset which gives maximum cross-correlation between $APF_{smooth}$ and $IPF_{smooth}$ after above preprocessing, is determined.
\end{description}

\begin{figure}[t]
	\centering
	\subfloat[]{\includegraphics[width=4.2cm]{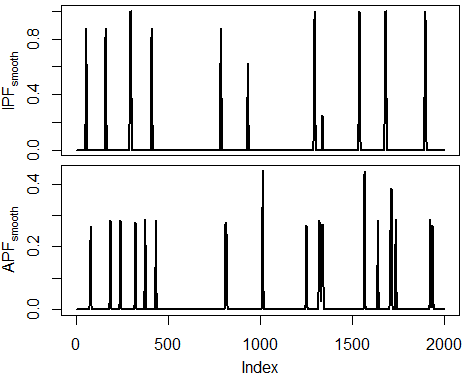}\label{image_peak_functions}}
	\quad
	\subfloat[]{\includegraphics[width = 4.0cm]{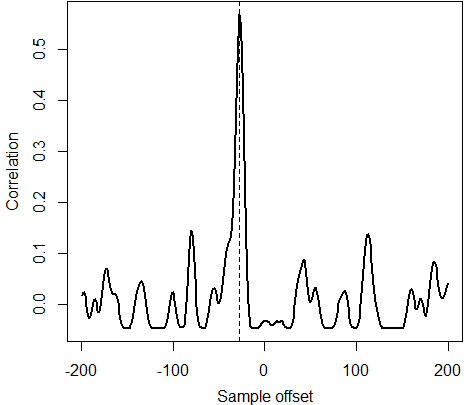}\label{image_correlation}}
	\quad
	\caption{\protect\subref{image_peak_functions} Plots of IPF and APF after density based binning and triangle smoothening and \protect\subref{image_correlation} cross-correlation plot between audio and IMU data}
	\label{fig:synchronize}
\end{figure}

Step 1 ensures that both APF and IPF signals are represented in a uniform scale for further steps. In Step 2, we try to penalize signal alignment based on the distance between peaks. Fig. \ref{fig:synchronize} shows the IPF and APF signals after Steps 1 and 2. Following Step 3 on these signals, the cross-correlation plot achieves a maxima at about -270ms, with a value of $0.6$, as shown in Fig. \ref{fig:synchronize}. Using the aforementioned synchronization procedure, it is seen that audio and IMU series can be aligned using a small snippet of sensor recording (20 seconds in Fig \ref{fig:synchronize}) with mean absolute error of $32ms$.

\section{Combined Shot Detector}
We use likelihood sequences from audio and IMU data (after synchronizing them) to detect shots. The algorithm is as follows:
\begin{enumerate}
	\item Maintain a buffer of input data from microphone and IMU. For every point in the IPF signal, if it is local maxima in its neighborhood of 500ms, choose the point as a candidate point.
	\item Maximum values of APF, IPF, $a_{rad}$, $a_{tan}$, $\omega_{rad}$ in the neighborhood of candidate points form the feature set to classify the region as shot or non-shot. Use pre-trained classifier to appropriately perform classification using the listed features. 
	\item If consecutive neighborhoods are labeled as shots, then keep only the first.
\end{enumerate}
The neighborhood region is so chosen that two governing peaks in IMU and audio will lie within this region. Also, it was observed that all five features considered above show a correlation of more than 0.4 with the output class (shot or not). Further, a suitable classification model is trained using these features to distinguish between a shot and non-shot.

\section{Results}
We used Samsung Gear S2 smartwatch to collect audio data at 8kHz and IMU data at 100Hz of $\sim650$ table tennis shots from a total of 8 players. The ranges of accelerometer and gyroscope sensors were $\pm8g$ and $\pm2000degrees/s$ respectively. The players (comprising of intermediate and professional players) were asked to play singles games after wearing Gear, while we video recorded them using Samsung S6 smartphone at 30fps. The videos assisted in manually tagging shots in audio and IMU data using an in-house developed tool. Entire dataset is divided into 80\% training and 20\% cross-validation and results are reported for cross-validation set.

We discuss the results of our developed Audio and IMU based shot detectors individually, the combined shot detector and how they compare with prior works. Table \ref{table:audio} shows that using a bandpass IIR filter (10 taps) obtained from conventional filter design techniques and APF, we obtain 75\% F-score for detection, which is better than using EPD-only block \cite{zhang2006ball}. Although, this method is computationally inexpensive, the accuracy saturates beyond 10 filter taps. With the proposed filter (23 taps) design based on the computational graph and APF, we obtain a higher F-score of 80\%, comparable to the EPD+MBR (C-SVC)\cite{zhang2006ball}. As shown in Table \ref{table:imu}, our IMU based shot detector gives an F-score of 78\%, performing better than Srivastava's Pan Tompkins's based detector \cite{srivastava2015efficient} which we implemented and tested on our dataset. Finally, our Combined shot detector obtains the best detection F-score of 95.6\%, when using a 50 trees random forest classifier as shown in Table \ref{table:combined}.


\begin{table}[!htb]
	\caption{Comparison of Audio based shot detector} 
	\vspace{-4pt}
	\centering 
	\begin{tabular}{cc c c c} 
		\hline\hline 
		Method & Precision & Recall & F-Score \\ [0.4ex] 
		\hline 
		EPD only \cite{zhang2006ball} & 28\%  & 84\% & 42\% \\ 
		EPD + MBR (C-SVC) \cite{zhang2006ball} & 91\%  & 73\% & 81\% \\ 
		IIR filter (10 taps) + APF & 65\% & 88\% & 75\%\\
		Trained filter (23 taps) + APF &85\% & 75\% & 80\%\\		
		\hline 
		\label{table:audio}
	\end{tabular}
	\caption{Comparison of IMU based shot detector} 
	\vspace{-4pt}
	\begin{tabular}{cc c c c} 
		\hline\hline 
		Method & Precision & Recall & F-Score \\ [0.5ex] 
		\hline 
		IMU Peak Function (IPF) & 73\% & 83\% & 78\%\\
		Srivastava \cite{srivastava2015efficient} on our dataset & 54\%  & 55\% & 55\% \\ 
		\hline 
		\label{table:imu}
	\end{tabular}
	\caption{Comparison of methods for our combined detector} 
	\vspace{-4pt}
	\begin{tabular}{cc c c c} 
		\hline\hline 
		Method & Precision & Recall & F-Score \\ [0.5ex] 
		\hline 
		SVM-RBF &88.4\% &94.8\% &91.5\% \\
		Random Forest & 97.3\%  & 93.9\% & 95.6\% \\ 
		\hline 
		\label{table:combined}
	\end{tabular}
\end{table}
\vspace{-28pt}
\section{Conclusion and Discussion}
We proposed a novel technique based on fusion of IMU and microphone data to detect shots in sports using wearables. The proposed Audio Peak Function (APF) with filter and IMU Peak Function (IPF) form the basis of shot detection using audio and IMU inputs respectively. We also discussed a methodology to synchronize the input streams in time, using peaks in APF and IPF signals. Specifically, for table tennis, our system achieves a high F-score of 95.6\% for shot detection, which is about 15\% improvement over using these sensors individually. Our proposed shot detector can form the basis of wearable-based analytics in sports e.g. table tennis, badminton etc. to calculate further features like shot-type distribution, speed, consistency, sweet spot analysis to help players gain deeper insights into their game.


\bibliographystyle{IEEEbib}
\bibliography{strings,refs}

\begin{thebibliography}{10}

\bibitem{pei2017embedded}
Weiping Pei, Jun Wang, Xubin Xu, Zhengwei Wu, and Xiaorong Du,
\newblock ``An embedded 6-axis sensor based recognition for tennis stroke,''
\newblock in {\em Consumer Electronics (ICCE), 2017 IEEE International
  Conference on}. IEEE, 2017, pp. 55--58.

\bibitem{srivastava2015efficient}
Rupika Srivastava, Ayush Patwari, Sunil Kumar, Gaurav Mishra, Laksmi
  Kaligounder, and Purnendu Sinha,
\newblock ``Efficient characterization of tennis shots and game analysis using
  wearable sensors data,''
\newblock in {\em SENSORS, 2015 IEEE}. IEEE, 2015, pp. 1--4.

\bibitem{connaghan2011multi}
Damien Connaghan, Phillip Kelly, Noel~E O'Connor, Mark Gaffney, Michael Walsh,
  and Cian O'Mathuna,
\newblock ``Multi-sensor classification of tennis strokes,''
\newblock in {\em Sensors, 2011 IEEE}. IEEE, 2011, pp. 1437--1440.

\bibitem{yoshikawa2010automated}
Fumito Yoshikawa, Takumi Kobayashi, Kenji Watanabe, and Nobuyuki Otsu,
\newblock ``Automated service scene detection for badminton game analysis using
  chlac and mra,''
\newblock {\em World Academy of Science, Engineering and Technology}, vol. 4,
  pp. 841--4, 2010.

\bibitem{wang2004sports}
Jinjun Wang, Changsheng Xu, Engsiong Chng, and Qi~Tian,
\newblock ``Sports highlight detection from keyword sequences using hmm,''
\newblock in {\em Multimedia and Expo, 2004. ICME'04. 2004 IEEE International
  Conference on}. IEEE, 2004, vol.~1, pp. 599--602.

\bibitem{zhang2006ball}
Bin Zhang, Weibei Dou, and Liming Chen,
\newblock ``Ball hit detection in table tennis games based on audio analysis,''
\newblock in {\em Pattern Recognition, 2006. ICPR 2006. 18th International
  Conference on}. IEEE, 2006, vol.~3, pp. 220--223.

\bibitem{xiong2003audio}
Ziyou Xiong, Regunathan Radhakrishnan, Ajay Divakaran, and Thomas~S Huang,
\newblock ``Audio events detection based highlights extraction from baseball,
  golf and soccer games in a unified framework,''
\newblock in {\em Acoustics, Speech, and Signal Processing, 2003.
  Proceedings.(ICASSP'03). 2003 IEEE International Conference on}. IEEE, 2003,
  vol.~5, pp. V--632.

\bibitem{rui2000automatically}
Yong Rui, Anoop Gupta, and Alex Acero,
\newblock ``Automatically extracting highlights for tv baseball programs,''
\newblock in {\em Proceedings of the eighth ACM international conference on
  Multimedia}. ACM, 2000, pp. 105--115.

\bibitem{manin2012vibro}
Lionel Manin, Florian Gabert, Marc Poggi, and Nicolas Havard,
\newblock ``Vibro-acoustic of table tennis rackets at ball impact: Influence of
  the blade plywood composition,''
\newblock {\em Procedia Engineering}, vol. 34, pp. 604--609, 2012.

\bibitem{kingma2014adam}
Diederik Kingma and Jimmy Ba,
\newblock ``Adam: A method for stochastic optimization,''
\newblock {\em arXiv preprint arXiv:1412.6980}, 2014.

\end{thebibliography}

\end{document}